\documentclass[runningheads]{llncs}
\usepackage[T1]{fontenc}
\usepackage{graphicx}
\usepackage{amsmath}
\usepackage{amssymb}
\usepackage{subfig}
\usepackage{booktabs}

\usepackage{algorithmicx}
\usepackage{algorithm}
\usepackage[noend]{algpseudocode}
\algrenewcommand{\Return}{\State\algorithmicreturn~}

\begin{document}
\title{Gradual Capacity Growth for Sparse Network Discovery}
\titlerunning{Gradual Capacity Growth for Sparse Network Discovery}
%
\author{Qihang Yao\inst{1}\orcidID{0000-0001-5091-3850} \and \\
Constantine Dovrolis\inst{1,2}\orcidID{0000-0002-4491-861X}}
\authorrunning{Q. Yao and C. Dovrolis}
%
\institute{Georgia Institute of Technology, Atlanta, GA, USA\\
\email{\{qihang,constantine\}@gatech.edu}
\and
The Cyprus Institute, Nicosia, Cyprus}
\maketitle              
\begin{abstract}
Sparse neural network methods typically assume that the target sparsity (or density) is fixed in advance, even though the relationship between network capacity and performance is generally unknown and task-dependent.
Existing approaches (including iterative pruning, dynamic sparse training, and pruning at initialization) either rely on dense pretraining, incur substantial retraining cost, or require a preset sparsity budget.
We propose Gradual Capacity Growth (GCG), a constructive sparse-to-dense training framework that allocates network capacity progressively during training.
Starting from a sparse seed network, GCG grows new connections in stages using PathGrow, a probabilistic path-based growth rule that biases additions toward high-signal input–output pathways while preserving structural diversity.
Growth is interleaved with limited training, and a lightweight performance–density extrapolation rule is used to estimate the smallest density beyond which further capacity increases yield diminishing accuracy gains. 
Experiments on CIFAR, TinyImageNet, and ImageNet show that GCG efficiently identifies sparse networks that achieve near-dense performance at moderate densities, without requiring dense pretraining or exhaustive retraining at each sparsity level.
While growth-only methods do not reach the extreme sparsity achievable by pruning or dynamic reallocation techniques, GCG substantially reduces total training cost compared to iterative magnitude pruning and provides a practical mechanism for exploring the accuracy–density tradeoff under limited optimization budgets.

\keywords{Sparse neural networks \and Density discovery \and Dynamic connectivity.}
\end{abstract}
\section{Introduction}
Artificial neural networks (ANNs) power state-of-the-art systems in vision, language, and many other domains. Their success has largely been driven by scaling: larger and denser models trained on larger datasets consistently improve performance.
Yet this progress comes at immense computational cost, motivating the search for sparse alternatives that retain dense-level accuracy with reduced training and inference requirements.

A central challenge in sparse learning is that, for a given task and architecture, the relationship between network density and performance is generally unknown.
Empirically, performance often saturates beyond a certain density, suggesting the existence of an \emph{operating density}—the smallest density at which near-dense performance is achieved.
Yet identifying this operating density typically requires expensive trial-and-error or dense pretraining, limiting both scientific understanding and practical deployment of sparse models.

The lottery ticket hypothesis (LTH)~\cite{frankle2018lottery} provided early evidence that dense networks contain sparse subnetworks (“winning tickets”) that can be trained in isolation to match full-model accuracy.
Iterative Magnitude Pruning (IMP) demonstrates such subnetworks, but at prohibitive cost—often 3--4$\times$ the FLOPs of dense training.
This has led to extensive work on pruning-at-initialization~\cite{lee2018snip}, dynamic sparse training~\cite{evci2020rigging}, and reparameterization-based sparsification~\cite{kusupati2020soft}.
Despite their differences, these approaches share a common assumption: the target density is specified in advance and remains fixed throughout training.

We argue that discovering the operating density is a central problem in sparse learning.
Rather than pruning a dense network or maintaining a fixed sparsity budget, we study a constructive sparse-to-dense approach in which growth searches over densities until performance saturates.

Our method, \emph{Gradual Capacity Growth (GCG)}, starts from a sparse seed and progressively adds connections using PathGrow, a probabilistic path-based growth rule inspired by path-kernel analysis.
Randomized growth helps avoid bottlenecks, while an exponential saturation rule estimates when additional density yields diminishing returns.
GCG therefore constructs a sparse subnetwork and estimates its operating density without dense pretraining or a preset sparsity budget.

\begin{figure}[t]
    \centering
    \includegraphics[width=\textwidth]{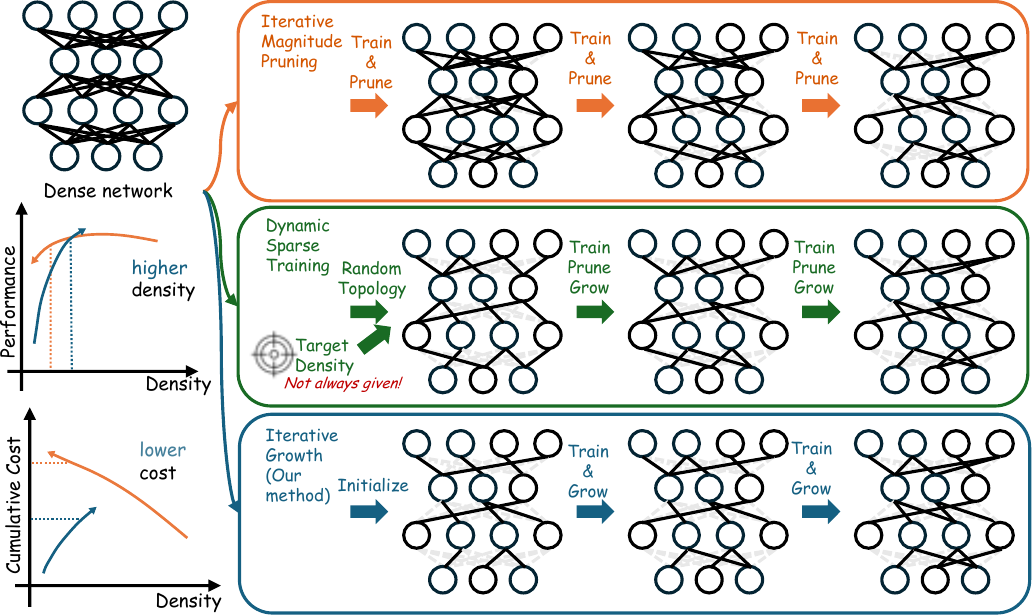}
    \caption{
        Comparison of pruning-, rewiring-, and growth-based strategies for sparse network discovery.
        The performance-density and cost-density sketches highlight the contrasting trade-offs between pruning and growth.}
    \label{fig:introduction-figure}
\end{figure}

As shown in Figure~\ref{fig:introduction-figure}, this growth-based view complements pruning and dynamic sparse training.
GCG typically reaches higher densities than IMP-derived tickets, but with lower cumulative training cost, making it a practical tool for exploring the moderate-sparsity regime.

\section{Problem Statement}
\label{sec:problem-statement}

Training sparse neural networks involves a trade-off between model density, task performance, and training cost.
For a given task and architecture, performance typically increases with density and then saturates, suggesting the existence of an \emph{operating density} at which near-dense performance is achieved.
However, this operating density is generally unknown in advance.

Let $T$ denote a learning task and $G$ a dense neural network with parameters $\theta \in \mathbb{R}^n$.
A sparse subnetwork $G'$ is defined by a binary mask $m \in \{0,1\}^n$ with effective parameters $\theta' = m \odot \theta$ and density $\rho(G') = \|m\|_0 / n$.
Let $P(G')$ and $C_{\mathrm{total}}(G')$ denote the performance and total training cost of $G'$, respectively.
Training cost is measured as cumulative floating-point operations (FLOPs).

Let $G_\rho$ denote the subnetwork returned by a procedure at density $\rho$.
Our objective is to identify the smallest density $\rho^*$ such that
\(
P(G_{\rho^*}) \ge P(G) - \delta,
\)
for a small tolerance $\delta$, while keeping $C_{\mathrm{total}}(G_{\rho^*})$ low.

IMP can approximate this operating density but requires multiple dense training cycles; instead, we seek a constructive procedure that grows a sparse network to discover $\rho^*$ without preset density.

\section{Related Works}
\label{sec:related-works}

\paragraph{Pruning-based Methods}
A common approach to sparse neural networks is pruning connections from a dense model.
Traditional pruning trains a dense network to convergence and removes low-magnitude weights, optionally followed by fine-tuning~\cite{hoefler2021sparsity}.
The Lottery Ticket Hypothesis (LTH)~\cite{frankle2018lottery} showed that dense networks contain sparse subnetworks (“winning tickets”) that can be trained from their original initialization to match dense accuracy, with Iterative Magnitude Pruning (IMP) providing empirical evidence.
To avoid the cost of dense pretraining, pruning-at-initialization methods such as SNIP~\cite{lee2018snip}, SynFlow~\cite{tanaka2020pruning}, and PHEW~\cite{patil2021phew} estimate parameter importance before training.
Recent work such as DPaI~\cite{xiang2025dpai} further optimizes the pruning mask at initialization using differentiable node–path balancing.
However, these approaches assume a target density that must be specified in advance.

\paragraph{Dynamic Sparse Training}
Dynamic Sparse Training (DST) maintains sparsity throughout training while periodically rewiring connections.
Typical methods alternate between pruning unimportant weights and adding new ones using random selection~\cite{mocanu2018scalable} or gradient-based criteria~\cite{evci2020rigging,dettmers2019sparse}.
Despite adapting topology during training, DST methods still require the target density to be fixed beforehand.

\paragraph{Growth-based Capacity Expansion}
Another line of work grows network capacity during training, typically by expanding width.
Methods such as Firefly~\cite{wu2020firefly} and GradMax~\cite{evci2022gradmax} progressively add neurons or channels to accelerate optimization.
These approaches usually converge to dense models, making sparsity temporary rather than a property of the final architecture.

\paragraph{Optimization-based Sparsification}
Optimization-based approaches treat the sparsity mask as a learnable parameter.
Methods starting from STR~\cite{kusupati2020soft} optimize sparsity directly during training, while other work optimizes binary masks with fixed random weights~\cite{ramanujan2020s}.
Similar to DST, these approaches adapt topology but still require a target density or sparsity constraint.

Unlike these methods, GCG does not assume a preset density and instead discovers the operating density through iterative growth during training.

\section{Method}
\label{sec:method}


\textbf{GCG} requires choices about initialization, growth timing, growth location, growth magnitude, weight initialization, and stopping. The core component is \textbf{PathGrow}, which determines where new connections are added.

\subsection{Where to Grow? Favor High-Weight Paths and Avoid Bottlenecks}


\paragraph{Create High-weight Paths for Faster Convergence}

Consider a feed-forward network $f(\cdot,\theta)$ trained by (stochastic) gradient descent.
One update step is given by
\begin{equation}
    f_{t+1}=f_t-\eta\,\Theta_t\,\nabla_f\mathcal L,
    \qquad
    \Theta_t=\nabla_\theta f_t\,
    \nabla_\theta f_t^{\!\top},
\end{equation}
with $\eta$ the learning rate, $\Theta_t$ the \emph{Neural Tangent Kernel} (NTK), and $ \mathcal{L} $ the empirical loss function.
$\nabla_f\mathcal L$ denotes the gradient of the loss with respect to the network outputs.
Directions whose NTK eigenvalues are large lead to faster convergence~\cite{arora2019fine}.
Gebhart \emph{et al.}~\cite{gebhart2021unified} express the network output at node $k$ as a sum over paths
\begin{equation}
    f^k(x,\theta)=\sum_{s}
    \sum_{p\in\mathcal P_{s\!\rightarrow k}}\pi_p(\theta)
    \,a_p(x,\theta)\,x_s,
\end{equation}
where $\mathcal P_{s\!\rightarrow k}$ is the set of paths from input $s$ to $k$,
$\pi_p(\theta)=\prod_{(i,j)\in p}\theta_{ij}$ is the path's weight product, and $a_p$ is an activation indicator.
Using the chain rule they factor the NTK into a data term and the \emph{Path Kernel} $\Pi_\theta=\nabla_\theta\pi(\theta)\,\nabla_\theta\pi(\theta)^{\!\top} $
which depends only on weights and topology.
Maximizing the trace of the path kernel accelerates convergence as it governs NTK eigenvalues.
With a newly added zero-weight-initialized connection $(i,j)$, the path kernel trace increases by
\begin{equation}
    \label{eq:delta-trace}
    \Delta\operatorname{Tr}(\Pi_\theta)_{(i,j)}=
    \sum_{p \mid (i,j)\in p} \Bigl( \prod_{(u, v) \in p'_{(i, j)}} \theta_{uv} \Bigr)^{\!2}.
\end{equation}
where $p'_{(i, j)} := p \setminus \{(i, j)\}$.
So, adding connections with high $\Delta\operatorname{Tr}(\Pi_\theta)_{(i,j)}$ is expected to speed up convergence.
$\Delta\operatorname{Tr}(\Pi_\theta)_{(i,j)}$ is expensive to compute, as it requires enumerating all paths in the network.
Instead of optimizing it explicitly, we introduce a $L_1$ surrogate that captures the dominant path-weight contributions while remaining efficient:
\begin{equation}
    \label{eq:pwmp-score}
    S(i, j) =
    \sum_{p \mid (i, j) \in p}
    \prod_{(u, v) \in p'_{(i, j)}}
    \lvert \theta_{u,v} \rvert
\end{equation}
We term this the \emph{Path Weight Magnitude Product} (PWMP) score, which is a heuristic growth signal motivated by NTK and path-kernel analysis.
PWMP serves as a computationally tractable proxy for prioritizing edges that participate in many high-magnitude paths.

\paragraph{Avoid Bottlenecks for Better Generalization}
While maximizing path kernel trace can speed up convergence, it does not guarantee good generalization.
Prior work~\cite{patil2021phew} shows that sparse networks that maximize trace under a fixed density tend to collapse into narrow hidden layers, or bottlenecks.
Such bottlenecks restrict the diversity of input-output paths and lead to worse generalization compared to broader, more evenly distributed structures.

Layer width alone does not capture bottlenecks, since connections can concentrate on a few nodes.
To quantify bottlenecks, we use the weighted $\tau$-core~\cite{batta2021weighted}, defined as the smallest set of nodes accounting for at least a fraction $\tau$ of total path centrality.
A small $\tau$-core indicates that paths concentrate through few nodes, while a larger $\tau$-core reflects more balanced path diversity.

\begin{figure}[t]
    \centering
    \includegraphics[width=\textwidth]{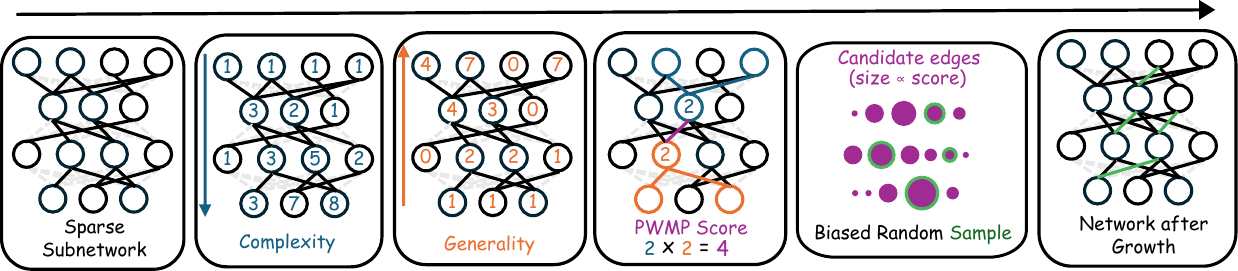}
    \caption{
        PathGrow computes PWMP scores by forward/backward passes and samples new connections proportionally to these scores.
    }
    \label{fig:pwmp_growth}
\end{figure}

\paragraph{Algorithm: PathGrow}
Based on the above insights, we propose a probabilistic growth algorithm, which we call PathGrow.
As illustrated in Figure~\ref{fig:pwmp_growth}, we first compute the PWMP score $S(i,j)$ for every potential connection $(i,j)$, and then sample new connections to add with probability proportional to $S(i,j)$.
By biasing growth toward high-PWMP edges while retaining randomness, PathGrow balances rapid convergence with structural diversity and mitigates bottlenecks.

$S(i, j)$ can be computed efficiently using a single forward and backward pass on a network where we convert every weight to its absolute value.
In the forward pass, we feed an all-ones input into the network.
The activation of each node $v$ is equal to the total PWMP for all paths from input to $v$, referred to as the complexity of $v$.
In the backward pass, we compute the gradient of the output with respect to node $v$'s pre-activation, which is equal to total PWMP for all paths from $v$ to output, referred to as the generality of $v$.
With these quantities, the PWMP gain $S(i,j)$ for a potential edge $(i, j)$ is computed as the product of the complexity of node $i$ and the generality of node $j$.
Both passes operate in  $\mathcal{O}(E)$ time, where $E$ is the existing edge count.
Pseudocode is provided in Appendix~\ref{sec:pseudocode}.

\subsection{Other Methodological Design Choices}

\paragraph{Where do we start? Initialize with PHEW}
We initialize the sparse network using PHEW~\cite{patil2021phew}, which provides a performant starting point at low density.
We choose the initial density $\rho_{init}$ to be low enough to allow thorough exploration of the density range, but high enough to avoid disconnected nodes (i.e. nodes with no incoming or outgoing connections).

\paragraph{When to Grow? Grow Early in the Training Process}

We grow early rather than after convergence.
Early training provides stronger and more coherent signals for selecting new connections, whereas later gradients are weaker and noisier~\cite{jastrzebski2020break,wang2020picking,dettmers2019sparse}.
We therefore interleave growth with short training phases, reducing cumulative training cost.
We refer to this short optimization period before each growth decision as \emph{rough training}: it is used only to expose useful path-weight signals and to update the performance--density trajectory.
In contrast, \emph{extensive training} denotes the longer post-growth training used only for final evaluation or ablation studies, after a candidate topology has been selected.
Thus, we adopt a training-and-growth strategy in which growth occurs after only a fraction $\omega$ of the total dense-network training budget, which reduces the cumulative training cost of the iterative process.
Empirically, we find that early growth yields final performance comparable to growth after longer training (Appendix~\ref{sec:growth-timing-result}).

\paragraph{When to Stop? Estimating the Operating Density}

A key challenge in growth-based training is determining when further increases in density no longer yield meaningful performance gains.
The goal is to estimate the \emph{operating density}—the smallest density at which performance saturates.

Because our method does not perform extensive training at every intermediate density, the final performance at each density is not directly observable.
We therefore use the performance--density trajectory observed during iterative growth as a proxy for identifying saturation.
Empirically, across datasets and architectures, performance exhibits diminishing returns as density increases, motivating a simple parametric model of this relationship.

To estimate the operating density, we fit an exponential saturation curve to the observed performance--density trajectory during growth:
\( P(G_k)=P_0+A(1-e^{-\beta\rho_k}) \).
We define \(\hat{\rho}^*\) as the smallest density at which the fitted curve reaches 95\% of the fitted improvement above \( P_0 \), i.e., \( P_0 + 0.95A \).

\paragraph{How Much to Grow? Exponential Density Growth}

At each iteration, we increase density by a fraction of the current density:
\(
    \Delta \rho_k = \gamma \cdot \rho_k,
\)
where $\gamma$ is a growth ratio.
This proportional strategy results in exponential growth, facilitating efficient exploration of the density-performance landscape.
We use $\gamma=25\%$ in our experiments, matching the pruning ratio of $20\%$ used in IMP~\cite{frankle2018lottery}.

\paragraph{How to Initialize New Connections? Zero Weights}

Following common practices in DST~\cite{evci2020rigging}, we initialize the weights of newly added connections to zero.
This avoids introducing noise or interference from random initializations and allows the network to learn appropriate values through gradient descent.
Zero initialization also maintains stability in the functional behavior of the current network.

\section{Experiments}
\label{sec:experiments}


\paragraph{Dataset and Architectures}
We evaluate our method on standard image classification tasks using \textit{CIFAR}, \textit{TinyImageNet} and \textit{ImageNet} to assess performance from small- to large-scale settings.
The corresponding network architectures include \textit{ResNet}~\cite{he_deep_2015} and Vision Transformer (\textit{ViT})~\cite{dosovitskiy_image_2021}.
For \textit{ViT}, we adopt scaled-down versions, following prior practices for training transformers on smaller datasets~\cite{lee2021vision}.

\paragraph{Pruning Scope}
For \textit{ResNet}, we follow the common setup~\cite{frankle2018lottery}, leaving the final linear layer and shortcut connections unpruned.
For \textit{ViT}, we follow the common practice of keeping the embedding layer and classification head dense, and additionally avoid pruning the query and key projection matrices in the attention modules.
Because path-kernel signals do not correlate well with Q/K weights, we keep those dense and defer attention-specific growth to future work.

\paragraph{Baseline Methods}
We have identified three categories of baseline methods for comparison, each serving a distinct purpose in evaluating our approach.
First, we compare against iterative magnitude pruning (IMP)~\cite{frankle2018lottery} in terms of performance, density and training cost.
We use the continued training variant (IMP-C) as our primary benchmark, since it produces stronger sparse networks by relaxing the requirement of retraining from scratch.
Second, we evaluate alternative growth strategies, including random growth~\cite{mocanu2018scalable} (RG) and gradient-based growth~\cite{evci2020rigging} (GG), to test the effectiveness of our method.
Third, we include methods that find good sparse networks given target densities, such as PHEW~\cite{patil2021phew}, RigL~\cite{evci2020rigging}, DSR~\cite{mostafa2019parameter}, SparseMomentum~\cite{dettmers2019sparse} and GSE~\cite{heddes2024always}.
While these methods do not directly address our problem of density discovery,
we compare performance of networks found at different density levels.

GCG trains a single model continuously as density increases, so parameters at higher densities inherit all earlier optimization.
In contrast, PHEW and RigL train separate models from scratch at each density.
To avoid penalizing these methods for not inheriting earlier training, we scale their training epochs so that the the cumulative trainable-parameter update budget (epochs × trainable parameters) matches matches that accumulated by GCG at the same density.
This normalization is not intended to model hardware runtime exactly; rather, it provides a topology-agnostic proxy for giving each method a comparable amount of optimization.

\section{Results}
\label{sec:results}

\paragraph{PathGrow Outperforms or Matches Other Growth Methods}

\begin{figure}[t]
  \centering
  \includegraphics[width=0.245\linewidth]{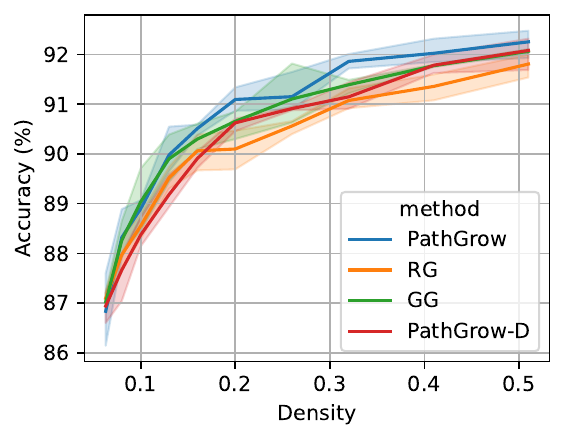}%
  \hfill
  \includegraphics[width=0.245\linewidth]{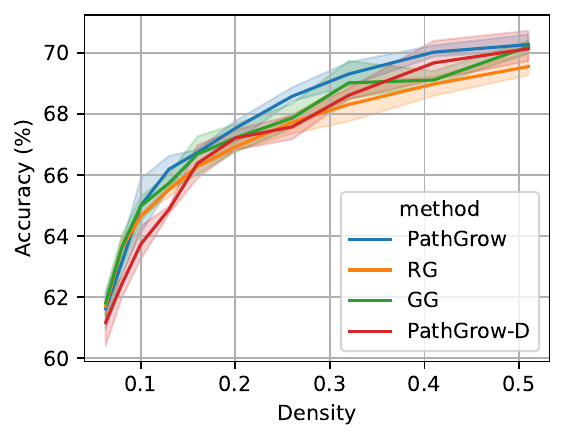}%
  \hfill
  \includegraphics[width=0.245\linewidth]{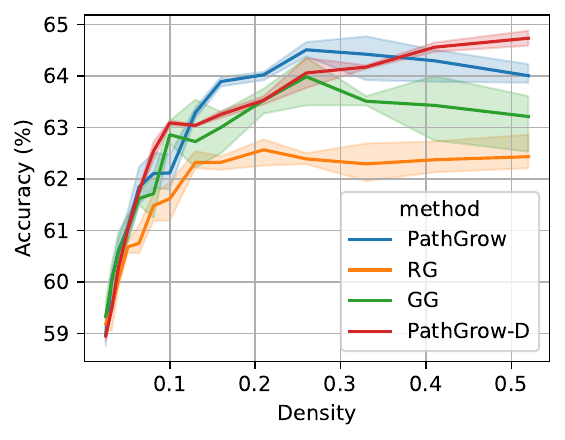}%
  \hfill
  \includegraphics[width=0.245\linewidth]{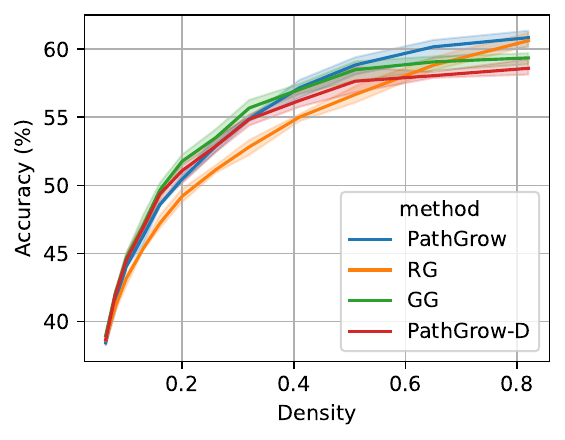}%
  \caption{Performance-density relationship of PathGrow compared with RG, GG and PathGrow-D. 
  (a) \textit{CIFAR-10}/\textit{ResNet-32}, (b) \textit{CIFAR-100}/\textit{ResNet-56}, (c) \textit{TinyImageNet}/\textit{ResNet-18}, (d) \textit{TinyImageNet} / \textit{ViT}.
  }
  \label{fig:compare-growth}
\end{figure}

Can PathGrow effectively identify strong sparse topologies?
We first compare it with gradient-based growth (GG) and random growth (RG), as shown in Figure~\ref{fig:compare-growth}.
On \textit{CIFAR} benchmarks, PathGrow matches or exceeds both baselines.
Its advantage becomes clearer on \textit{TinyImageNet / ResNet-18}.
For \textit{TinyImageNet / ViT}, PathGrow is slightly weaker than GG below  40\% density, but outperforms it at higher densities.
A key advantage of PathGrow is efficiency: GG estimates gradients of missing connections by temporarily treating the network as fully connected and processing a batch of examples, incurring high overhead, while PathGrow is purely topological and requires only one forward-backward pass on the sparse network.
We also assess an ablated variant, PathGrow-D, which deterministically adds connections that maximize total PWMP.
PathGrow-D generally underperforms PathGrow, supporting the importance of avoiding bottlenecks.
This is consistent with its favorable topological properties (Appendix~\ref{sec:appendix-topology}).

\paragraph{GCG Discovers Operating Densities with Near-Dense Performance}
\begin{figure}[t]
  \centering
  \includegraphics[width=0.24\linewidth]{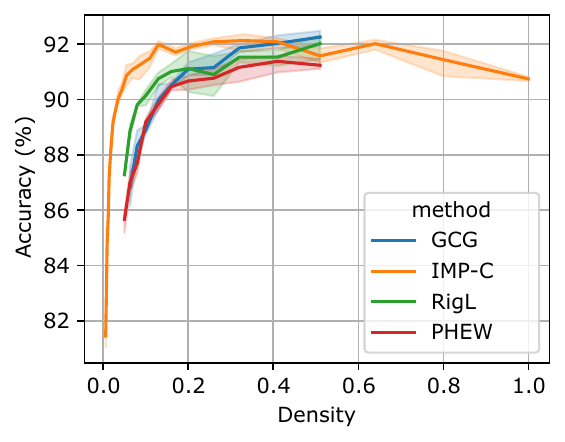}
  \hfill
  \includegraphics[width=0.24\linewidth]{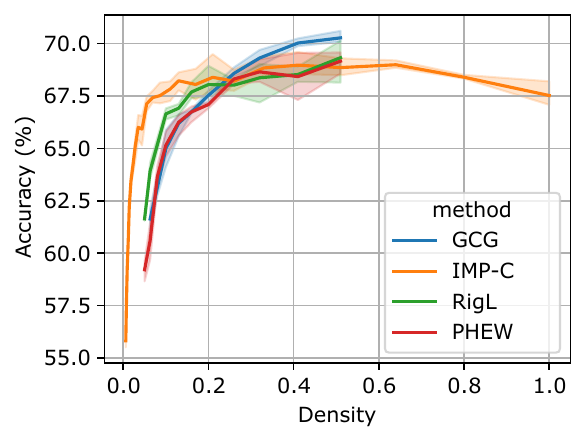}
  \hfill
  \includegraphics[width=0.24\linewidth]{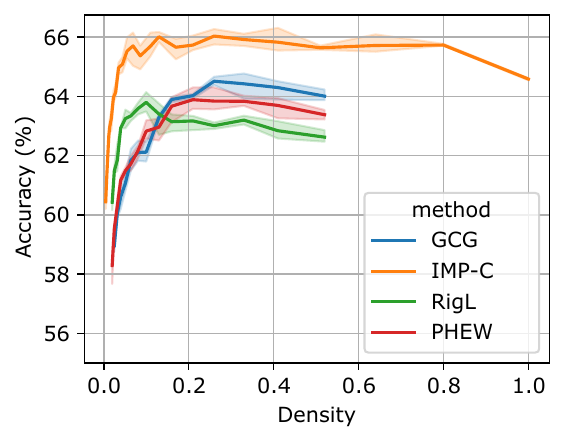}
  \hfill
  \includegraphics[width=0.24\linewidth]{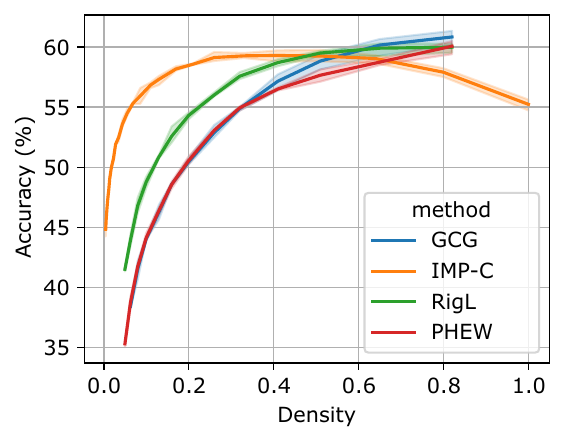}
  \caption{Performance–density curves comparing GCG with IMP-C, RigL, and PHEW. 
  (a) \textit{CIFAR-10}/\textit{ResNet-32}, (b) \textit{CIFAR-100}/\textit{ResNet-56}, (c) \textit{TinyImageNet}/\textit{ResNet-18}, (d) \textit{TinyImageNet} / \textit{ViT}.}
  \label{fig:performance-vs-density}
\end{figure}

Having established PathGrow as a strong growth rule within GCG, we now evaluate GCG’s ability to identify operating densities across datasets and architectures.
Figure~\ref{fig:performance-vs-density} shows performance-density tradeoffs across methods.
IMP-C was evaluated from 100\% to 1\% density, while GCG, PHEW, and RigL were run up to about 50\%; for PHEW and RigL, epochs were scaled to match GCG's cumulative training budget.

On \textit{CIFAR} benchmarks, GCG finds subnetworks that match or exceed IMP-C performance at densities of roughly 40\% and 30\%, respectively.
These densities are higher than those at which IMP-C retains near-optimal accuracy ($\sim$15\% and 20\%), reflecting IMP-C's advantage of pruning from a fully trained dense model, which provides richer importance signals.
On \textit{TinyImageNet / ViT}, it achieves parity but at around 50\% density—likely because the chosen \textit{ViT} architecture~\cite{heo2021rethinking,lee2021vision} is designed for small datasets and has substantially fewer parameters than \textit{ResNet-18}.
Figure~\ref{fig:performance-vs-density} also compares GCG with PHEW and RigL under matched training budgets.
GCG consistently outperforms PHEW, highlighting the benefit of iterative growth over one-shot pruning at initialization.
RigL is competitive at lower densities, but GCG surpasses it as density increases.

\begin{table}[t]
    \centering
    \caption{Top-1 accuracy on \textit{ImageNet} / \textit{ResNet-50}; GCG results averaged over 3 runs.}
    \label{tab:imagenet-result}
    \begin{tabular}{lcc}
        \toprule
        Method                                     & 10\% Density    & 20\% Density    \\
        \midrule
        DSR~\cite{mostafa2019parameter}           & 71.6            & 73.3            \\
        Sparse Momentum~\cite{dettmers2019sparse} & 72.3            & 73.8            \\
        RigL~\cite{evci2020rigging}               & 73.0 $\pm$ 0.04 & 75.1 $\pm$ 0.05 \\
        GSE~\cite{heddes2024always}               & 73.2 $\pm$ 0.07 & N/A             \\
        \textbf{GCG}                  & 71.0 $\pm$ 0.04 & 73.2 $\pm$ 0.13 \\
        \bottomrule
    \end{tabular}
\end{table}

In addition, we evaluate GCG on the large-scale \textit{ImageNet} dataset with \textit{ResNet-50}, to assess whether growth-based density discovery remains viable at scale.
Given the high computational cost, we report results at 10\% and 20\% density, comparing against dynamic sparsification methods.
GCG begins from a PHEW-initialized network at 2\% density and grows with a ratio of $\gamma=25\%$ until just below the target density, followed by a smaller one-step growth to match the target.
Table~\ref{tab:imagenet-result} summarizes the results: On ImageNet, GCG lags recent dynamic methods by about 2 percentage points, which we attribute to the disadvantage of growth-only training: unlike RigL or GSE, GCG cannot prune and reallocate connections that were added early but later become less useful. Nevertheless, these results show that growth-only density discovery remains viable at scale.

\paragraph{Iterative Growth is More Efficient than Iterative Pruning}

\begin{figure}[t]
    \centering
    \includegraphics[width=\linewidth]{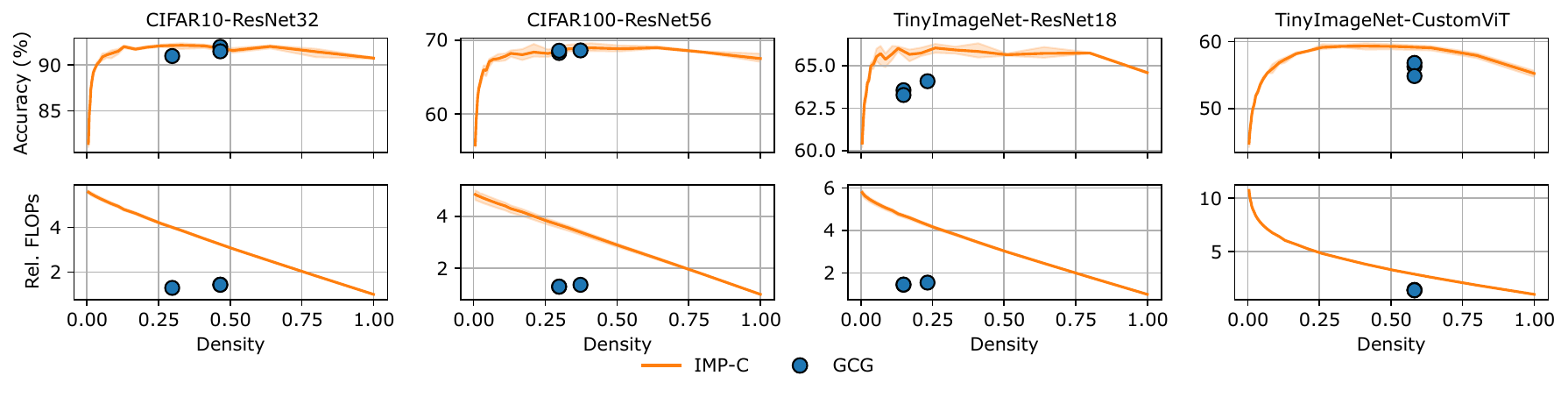}
    \caption{Accuracy and normalized cumulative cost of GCG and IMP-C across four dataset--architecture pairs.
        Top: Accuracy; Bottom: Relative FLOPs (normalized by extensive-dense training FLOPs).
        IMP-C curves represent results across all intermediate densities.
        GCG is shown as dots, with each dot representing one of three random seeds.
    }
    \label{fig:stopping-point-performance-cumulative-cost}
\end{figure}

Does growth-based density discovery require less training cost than pruning-based approaches?
We compare the performance and cumulative cost of GCG at the discovered operating density, with the performance and cost IMP-C achieves at all intermediate densities, as shown in Figure~\ref{fig:stopping-point-performance-cumulative-cost}.
This comparison is made at the operating density estimated by GCG, rather than at a density fixed in advance.
On both \textit{CIFAR} benchmarks, GCG matches IMP-C's performance while requiring only about $1.5\times$ the cost of training a dense model - less than half of the cost of IMP-C.
On \textit{TinyImageNet / ResNet-18}, GCG again reduces cost by more than half, though with a slight drop in accuracy relative to IMP-C.
On \textit{TinyImageNet / ViT}, the efficiency gain is smaller, likely because we use a compact ViT tailored for small datasets and the stopping density is higher ($\sim$58\%).
Overall, these results indicate that GCG efficiently discovers sparse networks whose accuracy is close to that achieved by IMP, but at substantially lower training cost.

\section{Discussion and Limitations}
\label{sec:conclusion}

We introduced GCG, a growth-based method for training sparse neural networks that simultaneously constructs subnetworks and discovers their operating density.
Unlike pruning approaches that destructively remove connections or dynamic sparse training methods that maintain a fixed sparsity level, GCG adopts a constructive sparse-to-dense paradigm: starting from a sparse seed, it grows edges using PathGrow, guided by path-weight signals, mitigates bottlenecks through randomized sampling, and stops when the fitted performance–density curve indicates diminishing returns.
This shift reframes sparse learning not only as an optimization problem but also as an exploration of how networks can grow into high-performing subnetworks.

Our results on CIFAR, TinyImageNet, and ImageNet demonstrate that GCG approaches the accuracy achieved by IMP, albeit at higher densities, and does so at substantially lower cost (~1.5x dense training FLOPs vs. 3-4x for IMP-C).
These findings establish growth-based density discovery as a complementary approach to pruning-based strategies for efficiently exploring the medium-sparsity regime in sparse training.

\paragraph{Limitations}
Despite these contributions, our work has several limitations.
First, growth-only methods cannot reach the extreme sparsity levels of pruning, since low-importance connections are never explicitly removed.
This explains why GCG requires higher density than IMP-C to match accuracy.
Second, the PWMP heuristic is tailored to feed-forward and convolutional structures; it does not naturally extend to query-key matrices in attention, where magnitudes decouple from functional importance.
Third, our efficiency analysis is in terms of algorithmic compute (FLOPs) rather than wall-clock time, and does not model hardware effects such as GPU kernel launch overheads, sparse memory access, or dynamic allocation costs, which may partially offset the gains of growth-based training on current accelerators.
Finally, our experiments are limited to vision benchmarks with relatively modest-scale transformers; validation in large-scale NLP or speech domains remains an important direction for future work.

Taken together, these limitations point to natural extensions:
hybrid grow-prune methods to combine the benefits of constructive and destructive updates;
attention-specific growth rules informed by synaptic diversity or head-level importance;
and broader domain validation.
More broadly, we view GCG as a step toward a broader growth-based research agenda for sparse learning — one that complements pruning and DST, and expands the conceptual landscape of how efficient subnetworks can be discovered.

\bibliographystyle{splncs04}
\bibliography{references}

\appendix

\section{PathGrow Pseudocode}
\label{sec:pseudocode}

\begin{algorithm}[H]
    \caption{PathGrow: Growing New Connections via Single-Pass PWMP Scoring}
    \label{alg:PWMPR-layer}
    \begin{algorithmic}[1]
        \Require Sparse network $G_k=(V,E_k)$ with weights $\theta$, growth budget $\Delta \rho(m_k)$, number of layers $L$
        \Ensure Updated network $G_{k+1}$ with $n \cdot \Delta \rho(m_k)$ new edges
        \State $n \leftarrow |\theta|$, \quad $M \leftarrow \lfloor n \cdot \Delta \rho(m_k) \rfloor$ \Comment number of edges to add
        \State $\tilde{\theta} \leftarrow |\theta|$ \Comment use absolute weights for PWMP computations

        \State \textbf{Forward pass (complexity).} Feed an all-ones input and propagate through the \emph{sparse} network using $\tilde{\theta}$ to obtain, at every pre-activation node $v$, its \emph{complexity} $\phi(v)$, i.e., the sum of absolute path-weight products from inputs to $v$.
        \State \textbf{Backward pass (generality).} Define a scalar readout by summing the final-layer pre-activations and backpropagate a unit signal through the \emph{sparse} network with $\tilde{\theta}$ to obtain, for each node $v$, its \emph{generality} $\psi(v)$, i.e., the sum of absolute path-weight products from $v$ to outputs.

        \For{$l=1$ to $L$}
        \For{each non-existent edge $(i,j)$ between layer-$l$ input node $i$ and output node $j$}
        \State $S(i,j) \leftarrow \phi(i) \cdot \psi(j)$ \Comment PWMP gain estimate for adding $(i,j)$
        \EndFor
        \EndFor

        \State Normalize scores over all missing edges to probabilities $P(i,j) \propto S(i,j)$.
        \State Sample $M$ edges without replacement from the set of missing edges according to $P(i,j)$.
        \State Add sampled edges to $E_k$ to obtain $E_{k+1}$ and \textbf{initialize all new weights to zero}.
        \Return $G_{k+1}=(V,E_{k+1})$
    \end{algorithmic}
\end{algorithm}

\section{Additional Experimental Analysis}

\subsection{Early Growth Creates Networks As Good As Growth After Convergence}
\label{sec:growth-timing-result}
\begin{figure}[h]
  \centering
  \includegraphics[width=0.49\linewidth]{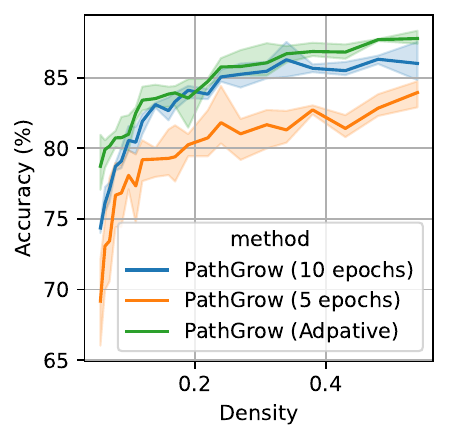}%
  \includegraphics[width=0.49\linewidth]{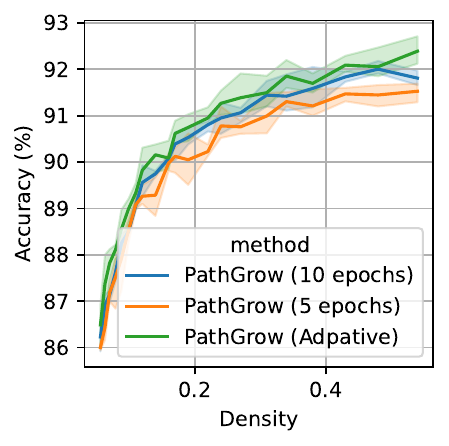}%
  \caption{Effect of growth timing on \textit{CIFAR-10} with \textit{ResNet-32}. We compare three schedules: 5 epochs, 10 epochs, and an adaptive early-stopping criterion at each density level. (a) Accuracy after rough training; (b) Accuracy after extensive training.}
  \label{fig:when-to-grow-example}
\end{figure}

How does the timing of growth decisions affect final performance?
We study this on \textit{CIFAR-10} and \textit{ResNet-32}.
In each iteration of GCG, we apply PathGrow, then perform a short period of training—“rough training”—before the next growth step.
We test rough training schedules of 5 epochs, 10 epochs, and an adaptive early-stopping rule that halts if validation loss fails to improve for 3 epochs.
After each density increment, we apply an “extensive training” phase to assess the resulting topology.
Figure~\ref{fig:when-to-grow-example} shows the results.
As expected, longer rough training improves intermediate accuracy.
However, after extensive training, performance differences become very small, especially between 10 epochs and the adaptive rule.
Thus, early growth—made after only a fraction~$\omega$ of full training—can produce topologies as effective as those grown after convergence.

\subsection{PathGrow Balances High Path Weight Magnitude Product and Low Bottlenecks}
\label{sec:appendix-topology}

\begin{figure}[h]
    \centering
    \includegraphics[width=\linewidth]{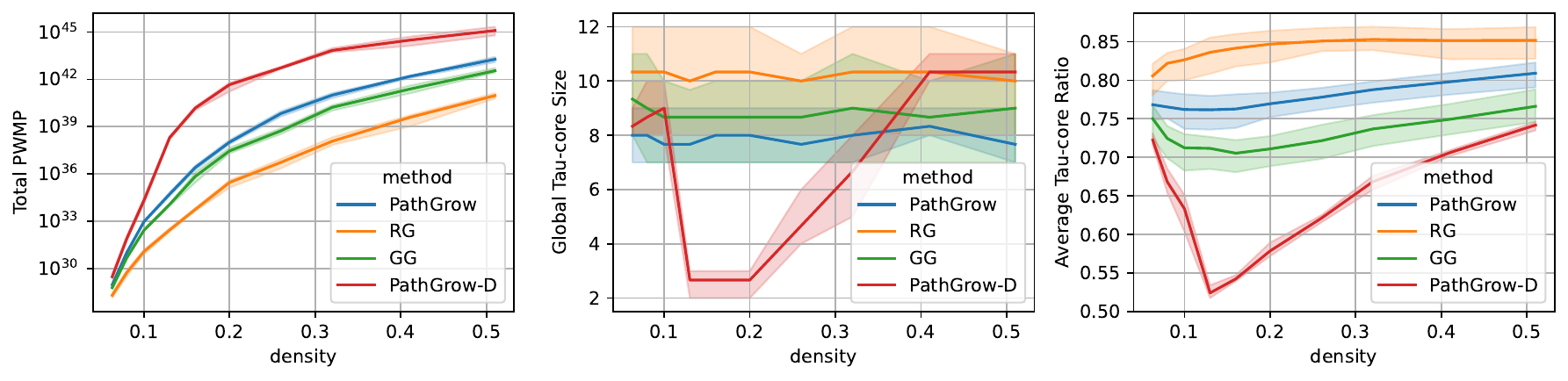}
    \caption{Evolution of topological metrics during iterative growth on \textit{CIFAR-10} / \textit{ResNet-32}, comparing different growth mechanisms (PathGrow, RG, GG and PathGrow-D).
        Left: Total PWMP.
        Middle: $\tau$-core size ($\tau=0.9$).
        Right: Average $\tau$-core ratio (higher = fewer bottlenecks).
    }
    \label{fig:topology}
\end{figure}
Do networks discovered by PathGrow exhibit the topological properties it is designed to encourage?
We evaluate two targeted metrics: the total PWMP and the $\tau$-core (with $\tau=0.9$).
Given the layered structure of neural networks, we measure the \textit{average $\tau$-core ratio} (layerwise $\tau$-core size normalized by width, then averaged) in addition to the global $\tau$-core size.
We analyze these metrics on \textit{CIFAR-10} with \textit{ResNet-32} as a representative case.

Figure~\ref{fig:topology} shows how these topological metrics evolve during growth under different strategies.
GG achieves a total PWMP comparable to PathGrow, suggesting that high-gradient connections also raise PWMP, but yields much lower average $\tau$-core ratios, indicating bottlenecks.
This may explain why PathGrow surpasses GG at higher densities, where generalization is critical.
RG yields the highest average $\tau$-core ratio, theoretically favoring generalization, but its low total PWMP—implying slow convergence—leads to worse performance.
Removing randomness from PathGrow boosts total PWMP but sharply lowers the average $\tau$-core ratio at low densities; beyond 13\%, the ratio recovers, likely because high-PWMP connections are already added and remaining edges distribute more evenly.

A final observation is that global $\tau$-core size is inversely correlated with performance, in contrast to the average $\tau$-core ratio.
We attribute this to residual connections in modern architectures such as \textit{ResNet}, which preserve effective pathways even when most connections in a block are pruned.
Thus, strong performance can persist despite a small global $\tau$-core, as long as some of the layers are still wide and remain accessible through residual links.

\end{document}